\documentclass{article}

\usepackage{PRIMEarxiv}

\usepackage[utf8]{inputenc} % allow utf-8 input
\usepackage{amsmath}
\usepackage{tabularx}
\usepackage[T1]{fontenc}    % use 8-bit T1 fonts
\usepackage{hyperref}       % hyperlinks
\usepackage{url}            % simple URL typesetting
\usepackage{booktabs}       % professional-quality tables
\usepackage{amsfonts}       % blackboard math symbols
\usepackage{nicefrac}       % compact symbols for 1/2, etc.
\usepackage{microtype}      % microtypography
\usepackage{lipsum}
\usepackage{fancyhdr}       % header
\usepackage{graphicx}       % graphics
\graphicspath{{media/}}     % organize your images and other figures under media/ folder

\usepackage{multirow}
\usepackage{makecell}
\title{Training Large-Scale Optical Neural Networks with Two-Pass Forward Propagation}

\author{
  Amirreza Ahmadnejad \\
Department of Electrical Engineering \\
	Sharif University of Technology\\
	Tehran, Iran \\
  \texttt{amirreza.ahmadnejad@sharif.edu } \\
     \And
  Somayyeh Koohi\\
  Department of Computer Engineering \\
	Sharif University of Technology\\
	Tehran, Iran \\
  \texttt{koohi@sharif.edu } \\
}

\begin{document}
\maketitle

\begin{abstract} 
This paper addresses the limitations in Optical Neural Networks (ONNs) related to training efficiency, nonlinear function implementation, and large input data processing. We introduce Two-Pass Forward Propagation, a novel training method that avoids specific nonlinear activation functions by modulating and re-entering error with random noise. Additionally, we propose a new way to implement convolutional neural networks using simple neural networks in integrated optical systems. Theoretical foundations and numerical results demonstrate significant improvements in training speed, energy efficiency, and scalability, advancing the potential of optical computing for complex data tasks.
\end{abstract}

%%%%%%%%%%%%%%%%%%%%%%%%%%  body  %%%%%%%%%%%%%%%%%%%%%%%%%%

\section{Introduction}
The challenge of unconventional various type data processing is crucial today, driven by the advanced capabilities of processors and the increasing complexity of datasets \cite{r1,r2,r17}. In this context, neural network architectures are essential for processing large datasets, inspired by biological neural networks. Despite significant progress, two major issues remain: handling extremely large datasets and training these models effectively. While models like Large Language Models \cite{r3} have made strides in addressing the first issue, challenges such as processing time, power consumption, and the overhead of implementing these models on electronic systems persist. This necessitates further research and innovation \cite{r4}.

One way to address these challenges is through optical implementations of neural networks. Leveraging light's unique properties and parallel processing capabilities, optical implementations promise widespread use with much lower power consumption. However, two main challenges remain: handling real-size data in an integrated format and training these optical networks. Progress has been made in processing large datasets, especially with free-space optical implementations \cite{r5,r6}. Still, a key issue is designing a chip that can match the processing power of nanometer-scale electronic processors. This goal requires a critical look at the fundamental objectives and feasibility of achieving comparable performance with optical neural networks.

Proposed solutions for training ONNs often adapt traditional backpropagation methods or integrate other techniques specific to electromagnetic physics. A major challenge in integrated implementation is the activation function. Unlike linear optical components with bidirectional behavior, creating a nonlinear block with the right characteristics is difficult. This challenge involves developing a nonlinear optical function \cite{r7}. Approaches inspired by the adjoint method combined with light physics principles have produced successful training algorithms \cite{r8}. However, they haven't fully addressed incorporating nonlinear functions during training. Hinton's novel Forward-Forward algorithm \cite{r9}, praised for its suitability for optical implementation, still faces issues with nonlinear function representation \cite{r10}. Even with this algorithm, designing a nonlinear function, known as a Nonlinear Transformer, remains necessary for effective training.

Another fundamental challenge is processing real-size datasets in integrated ONNs, regardless of the training methods used \cite{r11,r12}. Although significant progress has been made in miniaturizing these processors, most models struggle to process real-world data, unlike their electronic counterparts. This limitation prevents integrated ONNs from reaching their full potential. Innovative solutions are needed to enable ONNs to handle and process real-world visual data as effectively as electronic systems.

In this paper, we propose two independent solutions to enhance ONNs, addressing both the challenges of training and data processing:

\begin{itemize}
    \item \textbf{Two-Pass Forward Propagation Training Approaches}: 
    \begin{itemize}
        \item Develop a robust training method inspired by Backpropagation, Forward-Forward, and Adjoint Methods \cite{r7,r8,r9,r10}.
        \item Implement a novel error modulation approach where the error is re-entered into the network along the primary pathway, rather than from the output, simplifying the training process.

    \end{itemize}
    \item \textbf{Real Size Image Processing For Integrated ONNs}: 
    \begin{itemize}
        \item Propose a new method for implementing CNNs using simple feed-forward neural networks to overcome practical hurdles in optical integrated implementations.
        \item Form a comprehensive three-dimensional neural network capable of processing real-size images, enabling ONNs to rival electronic CNNs in handling large datasets effectively.
    \end{itemize}
\end{itemize}

These innovations aim to overcome current limitations and advance the capabilities of optical neural networks, making them more practical and efficient for real-world applications.

The structure of this article unfolds as follows:
In the second section, our proposed method for training ONNs is presented using the notation introduced in \cite{r8}. This section first discusses the mathematical model based on the derivative of the cost function with respect to the permittivity of phase shifters in the Mach-Zehnder mesh and the incorporation of variables from the electromagnetic fields. A comparison is then made between the existing training models for optical implementations, namely Backpropagation, Forward-Forward, and our proposed model.
In the third section, we present our main idea for achieving CNN equivalence and processing real-size data in an optical integrated mode.
In the fourth section, we discuss the numerical simulations performed to calculate the gradient and the complete training of a simple ONN with four inputs and four outputs (equivalent to an XOR gate) and a convolutional ONN using the idea from the third section. A comparison with existing benchmarks is also provided. The conclusion and future works are stated in the final section. The code for the simulations described in this paper is available at our GitHub repository\cite{r16}.

\section{Two-Pass Forward Propagation Training Approaches}
Our approach introduces an algorithm that initially performs similarly to backpropagation (BP) in its first phase. After the error calculation, it modulates this error and reintroduces it as an input to the model. The decision to modulate the input with the error signal is inspired by biological principles observed in neural processes within the brain. Specifically, the central philosophy is to integrate the error information directly into neural activities, rather than propagating the error through a separate backward pass\cite{r13}.

In biological systems, global neuromodulatory signals, such as those from neuromodulatory systems, influence activity at early stages of the visual processing hierarchy, including the primary visual cortex (V1)\cite{r13}. These signals are believed to convey information related to task performance or errors in the system's output. By directly incorporating error information into the input signal, the proposed method aims to emulate this biological mechanism of error-driven modulation of neural activities. This approach contrasts with traditional backpropagation, which requires a separate backward pass to propagate error gradients—an aspect considered biologically unrealistic. The input modulation in this method allows error information to be integrated into the forward propagation, thereby eliminating the need for a backward pass\cite{r13}.

\subsection{Mathematical Representation}
Inspired by the notation in the literature, we start with the forward pass for layers \( l = 1 \) to \( L \):
\begin{align}
\mathbf{Z}_l &= \hat{\mathbf{W}}_l \mathbf{X}_{l-1} \\
\mathbf{X}_l &= f_l(\mathbf{Z}_l)
\end{align}
Here, \(\mathbf{Z}_l\) is the pre-activation at layer \( l \), \(\hat{\mathbf{W}}_l\) is the weight matrix at layer \( l \), \(\mathbf{X}_{l-1}\) is the activation from the previous layer, \(\mathbf{X}_l\) is the activation at layer \( l \), and \(f_l\) is the activation function at layer \( l \).

After calculating the output, we can measure the error using specific metrics like mean squared error or others:
\begin{align}
\mathbf{\Gamma}_L &= \mathbf{X}_L - \mathbf{T}
\end{align}
where \(\mathbf{\Gamma}_L\) is the error at the output layer, \(\mathbf{X}_L\) is the network output, and \(\mathbf{T}\) is the target output.

We then modulate this error using a simple function which introduces random noise. This modulated signal re-enters the first layer. If the modulating function is denoted as \(\mathbf{F}(\cdot)\), we have:
\begin{align}
\mathbf{X}_{\text{err},0} &= \mathbf{X}_0 + \mathbf{F}(\mathbf{\Gamma}_L) \\
\mathbf{Z}_{\text{err},l} &= \hat{\mathbf{W}}_l \mathbf{X}_{\text{err},l-1} \\
\mathbf{X}_{\text{err},l} &= f_l(\mathbf{Z}_{\text{err},l}) \quad \text{for } l = 1 \text{ to } L
\end{align}
By this modulation (which in fact, as will be pointed out later, will actually be the numerical multiplication of the error $\mathbf{\Gamma}_L$ in the random matrix $\mathbf{F}$), the weights in the layers \( l=1 \) to \( L \) can be updated as follows:
\begin{align}
\Delta \hat{\mathbf{W}}_1 &= (\mathbf{X}_1 - \mathbf{X}_{\text{err},1}) \cdot (\mathbf{X}_0 + \mathbf{F}(\mathbf{\Gamma}_L))^T \\
\Delta \hat{\mathbf{W}}_l &= (\mathbf{X}_l - \mathbf{X}_{\text{err},l}) \cdot \mathbf{X}_{\text{err},l-1}^T \quad \text{for } l = 2 \text{ to } L-1 \\
\Delta \hat{\mathbf{W}}_L &= \mathbf{\Gamma}_L \cdot \mathbf{X}_{\text{err},L-1}^T
\end{align}
The weight updating rule is then inspired by previous methods as follows:
\begin{align}
\hat{\mathbf{W}}_l(t+1) &= \hat{\mathbf{W}}_l(t) - \eta \Delta \hat{\mathbf{W}}_l
\end{align}
where \(\eta\) is the learning rate and \(t\) represents the iteration.

\begin{figure}[htbp]  % Specify the placement of the figure (here: 'here', 'top', 'bottom', 'page')
    \centering
    \includegraphics[width=\textwidth, height=4cm]{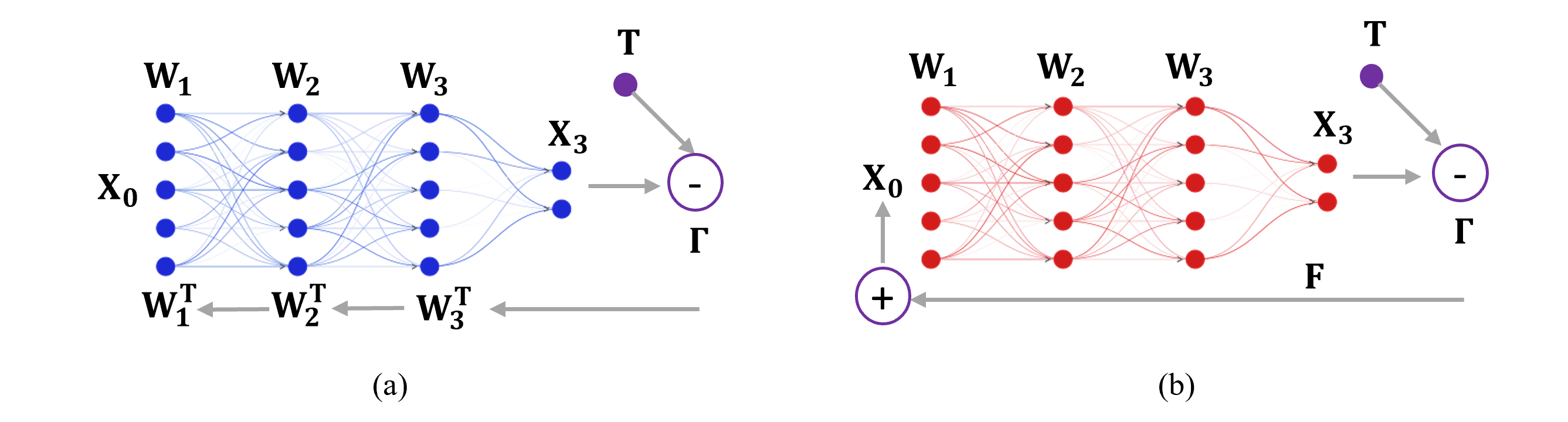}
    \caption{Comparison between the concept of (a)BackPropagation/Adjoint and (b) Two Forward Propagation training methods.}
    \label{fig1}
\end{figure}

A comparison between the backpropagation (Adjoint) training method and the two-pass forward propagation method is illustrated in Figure \ref{fig1}.

\subsection{Equating the algorithm with optical interpretations}\label{part2}
To represent the optical format, consider a simple feedforward \( L \)-layer neural network where \( \epsilon_l \) represents the permittivity of the weights in each layer. In the forward pass, for layers \( l = 1 \) to \( L \):
\begin{align}
\hat{\mathbf{A}}_l \mathbf{e}_l &= \mathbf{b}_{l-1} \\
\mathbf{X}_l &= f_l(\mathbf{P}_{\text{out},l} \mathbf{e}_l)
\end{align}
where $\hat{\mathbf{A}}_l$ is the Maxwell operator for layer \( l \), $\mathbf{e}_l$ is the electric field in layer \( l \), $\mathbf{b}_{l-1}$ is the source term from the previous layer,and $\mathbf{P}_{\text{out},l}$ extracts the output field amplitudes.
We then measure the intensity of the output ports. By comparing the measured value with the target ones, we calculate the error similar to Eq.3:
\begin{align}
\mathbf{\Gamma}_L &= \mathbf{X}_L - \mathbf{T}
\end{align}
Next, by modulating the error using a random behavior function, we re-enter these new signals into the network. The properties of weights and inputs are arranged as follows:
\begin{align}
\mathbf{b}_{\text{err},0} &= \mathbf{b}_0 + \mathbf{F}\times \mathbf{\Gamma}_L \\
\hat{\mathbf{A}}_l \mathbf{e}_{\text{err},l} &= \mathbf{b}_{\text{err},l-1} \\
\mathbf{X}_{\text{err},l} &= f_l(\mathbf{P}_{\text{out},l} \mathbf{e}_{\text{err},l}) \quad \text{for } l = 1 \text{ to } L
\end{align}
As mentioned, the purpose of random matrix $\mathbf{F}$ is to multiply with vector $\mathbf{\Gamma}_L$. The weights are updated by saving the values of permittivity in the first propagation. Then, by forward propagating the modulated error, we can update the weight values in the MZI phase shifters:
\begin{align}
\Delta \epsilon_1 &= k_0^2 \mathcal{R}\left\{\sum_{\mathbf{r} \in \mathbf{r}_\phi} (\mathbf{e}_1(\mathbf{r}) - \mathbf{e}_{\text{err},1}(\mathbf{r})) \cdot (\mathbf{e}_0(\mathbf{r}) + \mathbf{e}_F(\mathbf{r}))^* \right\} \\
\Delta \epsilon_l &= k_0^2 \mathcal{R}\left\{\sum_{\mathbf{r} \in \mathbf{r}_\phi} (\mathbf{e}_l(\mathbf{r}) - \mathbf{e}_{\text{err},l}(\mathbf{r})) \cdot \mathbf{e}_{\text{err},l-1}(\mathbf{r})^* \right\} \quad \text{for } l = 2 \text{ to } L-1 \\
\Delta \epsilon_L &= k_0^2 \mathcal{R}\left\{\sum_{\mathbf{r} \in \mathbf{r}_\phi} \mathbf{\Gamma}_L \cdot \mathbf{e}_{\text{err},L-1}(\mathbf{r})^* \right\}
\end{align}
where $k_0$ is the wavenumber, $\mathbf{r}_\phi$ represents the positions of the phase shifters, $\mathbf{e}_F = \mathbf{F}\times \mathbf{\Gamma}_L$ is the field corresponding to the error feedback.
The permittivity updating rule, related to changing the voltage over the materials of the phase shifters, is as follows:
\begin{align}
\epsilon_l(t+1) &= \epsilon_l(t) - \eta \Delta \epsilon_l
\end{align}
In this formulation, the forward and modulated passes are expressed in terms of solving Maxwell's equations for each layer. The permittivity updates are computed as overlap integrals of the difference between the original and modulated fields, similar to the adjoint method in the literature. The error feedback is incorporated into the source term for the modulated pass.

\begin{figure}[htbp]  % Specify the placement of the figure (here: 'here', 'top', 'bottom', 'page')
    \centering
    \includegraphics[width=12cm, height=4cm]{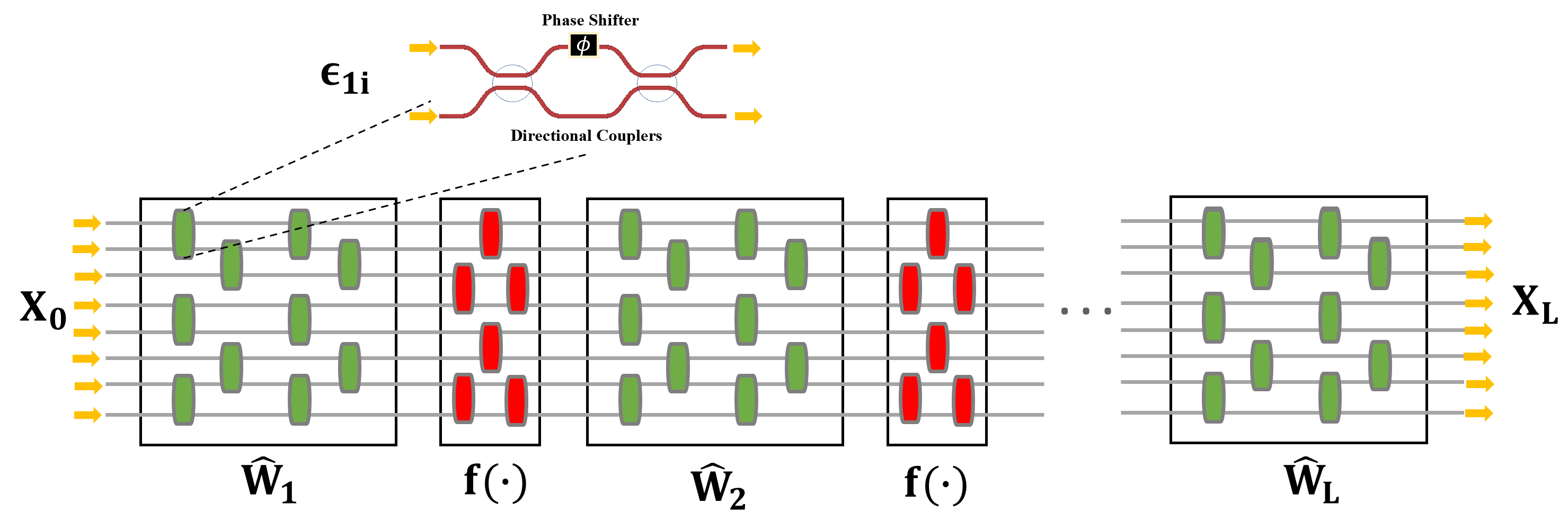}
    \caption{The outline of the implementation of the optical neural network using MZI mesh. Each of the modulators are stimulated electro-optically with a special voltage. The nonlinear block related to the activator function is also implemented using the MZI mesh and the reason for the color difference between them is the different material used as a phase changer in the modulator.}
    \label{fig2}
\end{figure}

A noteworthy point is that in this circuit, unlike traditional systems that convert light into voltage and then apply nonlinear operations, an electro-optical MZI mesh can be used to perform nonlinear operations directly\cite{r14}. The general structure of a simple ONN implementation using an MZI mesh for linear and nonlinear operations is shown in Figure \ref{fig2}.

\section{Real Size Image Processing}\label{part3}

\begin{figure}[htbp]  % Specify the placement of the figure (here: 'here', 'top', 'bottom', 'page')
    \centering
    \includegraphics[width=12cm, height=7cm]{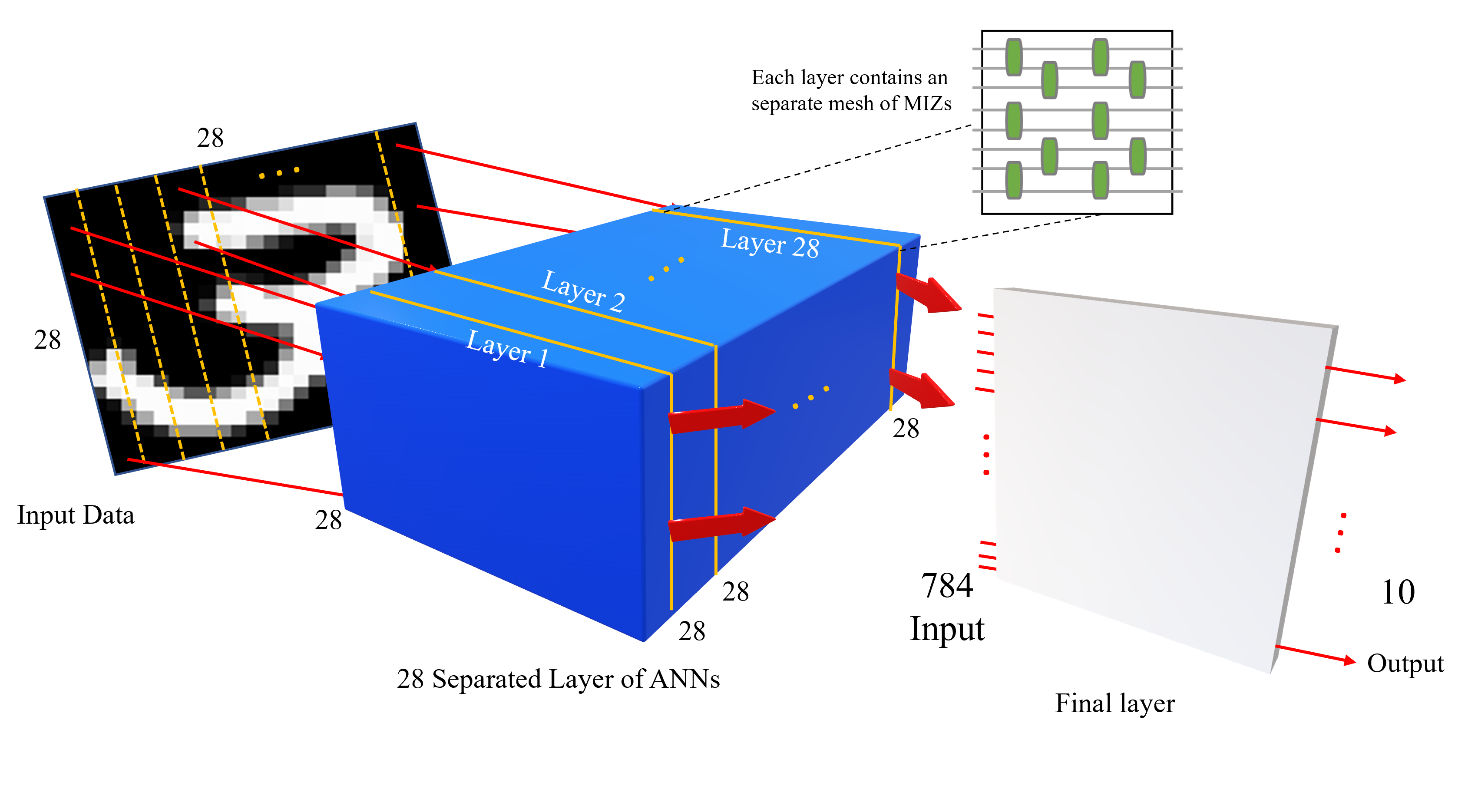}
    \caption{Scheme of converting convolutional neural network to simple neural networks.}
    \label{fig3}
\end{figure}

The training process of the new Optical Neural Network (ONN) involves splitting the input image into rows, processing each row through separate artificial neural networks (ANNs), and updating weights based on the computed errors. The general outline of this idea is shown in Figure \ref{fig3}. 
As illustrated in the figure, the input image, typically $28\times 28$ pixels and in black and white, is separated into columns (though separation can also be row-wise, simulations indicate that column-wise separation yields better results). Each column is then processed by a simple neural network independently. For a $28\times 28$ input image, each column results in an input to these networks comprising 28 input neurons. After performing the linear operation and applying the activation function, each network produces 28 output neurons. This number was found to be optimal in simulations but can be adjusted.

Subsequently, all these 28 outputs from each of the 28 networks (totaling 784 outputs) are fed into another network. This subsequent network has 28 input neurons arranged in three dimensions (the exact number can vary depending on the configuration of the previous layer). The final output consists of 10 neurons, corresponding to the number of classes in the MNIST dataset.

\begin{figure}[htbp]  % Specify the placement of the figure (here: 'here', 'top', 'bottom', 'page')
    \centering
    \includegraphics[width=13cm , height=6cm]{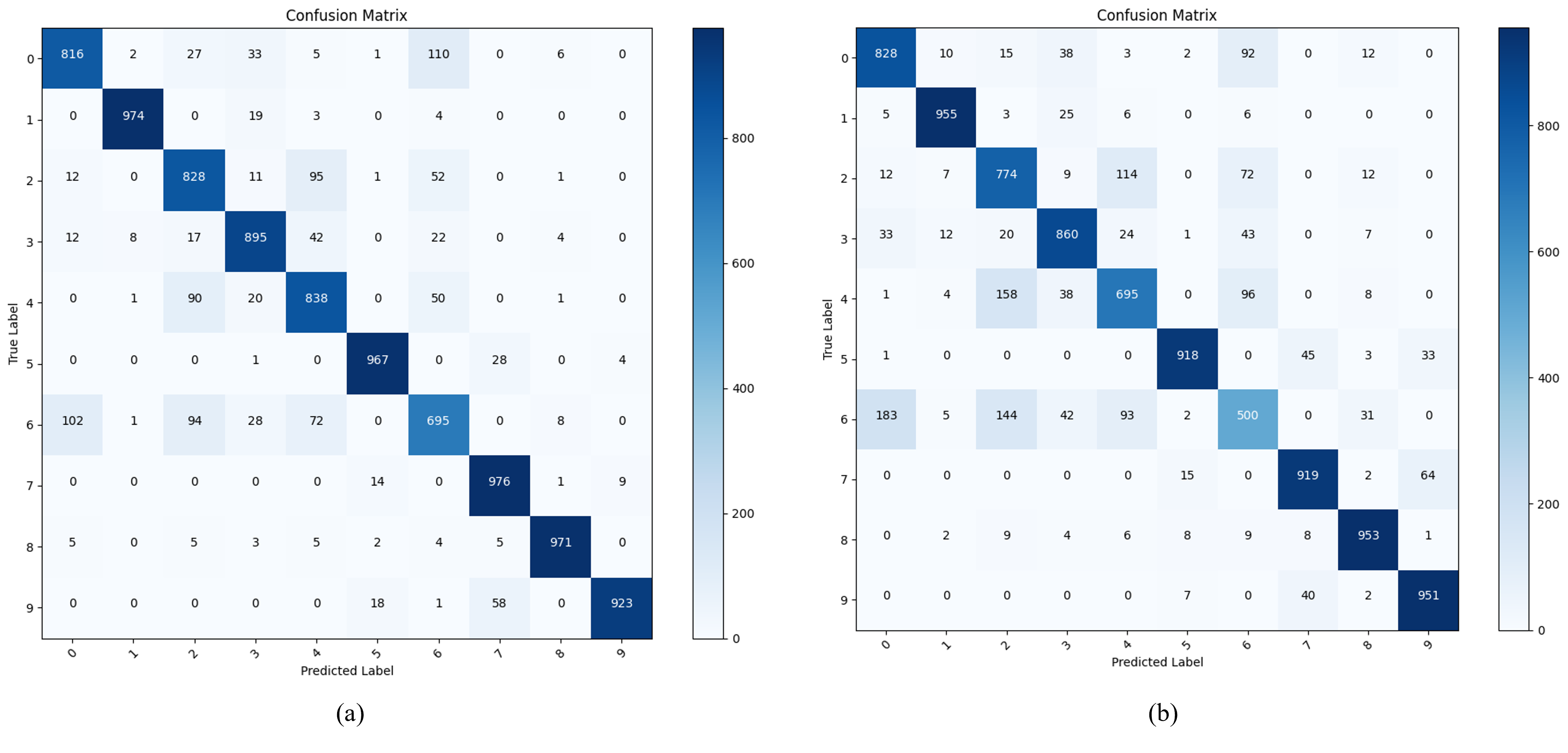}
    \caption{Equivalence of a convolutional network with a number of simple neural networks.}
    \label{fig4}
\end{figure}

To verify the correctness of this method, it was first implemented on a silicon system, in addition to the optical implementation. The confusion matrix of this network, compared to that of a 3-layer convolutional network, is shown in Figure \ref{fig4}.

From the comparison of these two models for the classification of MNIST, it is clear that our model can achieve acceptable results on silicon systems. This provides hope that it can also be effectively used for integrated optical convolutional neural networks.

\section{Numerical Implementation}
All simulations conducted on the structures mentioned in Lumerical are based on the Finite-Difference Time-Domain (FDTD) method. Inspired by the methodology described in \cite{r8}, we evaluate our Optical Neural Network (ONN) and the corresponding training procedure as an Optical XOR gate and a real integrated ONN. We then compare the final classification results on the MNIST dataset with those from previous well-known models.

\begin{figure}[htbp]  % Specify the placement of the figure (here: 'here', 'top', 'bottom', 'page')
    \centering
    \includegraphics[width=\textwidth , height=6cm]{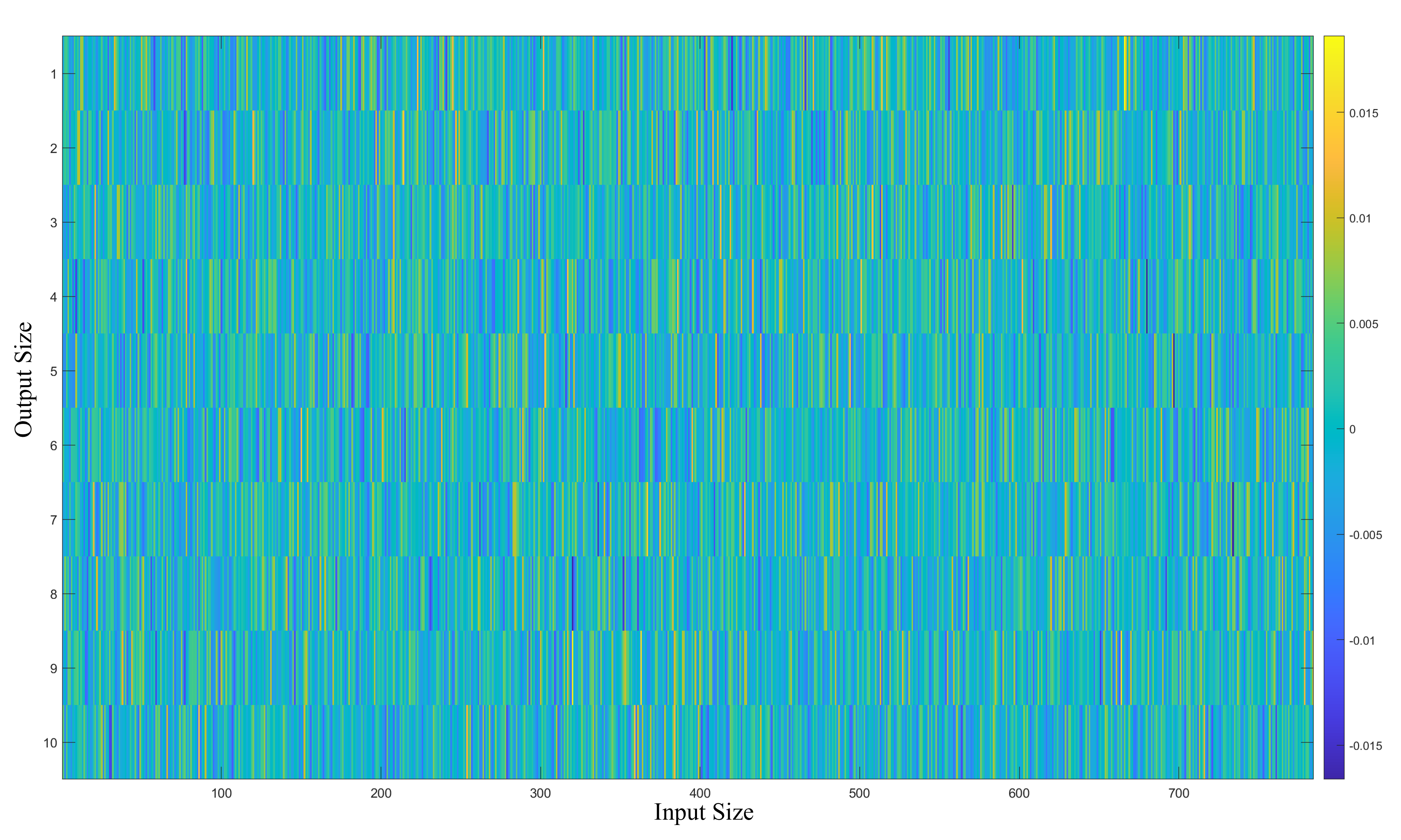}
    \caption{Visulation of random matrix $\mathbf{F}^T$ for 10 outputs and 784 inputs.}
    \label{fig5}
\end{figure}

A crucial aspect that must be highlighted is the role of the random \( \mathbf{F} \) matrix. As discussed in Section \ref{part2}, this matrix is initialized with elements drawn from a Gaussian (normal) distribution, with a mean of 0. The standard deviation of this distribution is calculated as:
\begin{equation}
    \sigma= 0.05 \times \sqrt{\frac{6}{F_{in}}}
\end{equation}
 where $F_{in}$ represents the input dimensionality or Fan-in (784 for MNIST). Thus, for MNIST, the standard deviation is approximately 0.004374. The variance is obtained by squaring the standard deviation, yielding $1.9132 \times 10^{-5}$.
 After calculating the error, the random matrix \( \mathbf{F} \) is scaled according to the output size of the neural network, combined with the input, and fed back into the model. An illustrative example of this process is depicted in Figure \ref{fig5} , where the ONN features 10 outputs and 784 ($28\times 28$) inputs.

\subsection{XOR Gate}
Using the method described in \cite{r8}, we first train a simple ONN to perform a task equivalent to an XOR gate. This gate has two inputs and one output: if both inputs are the same (both 0 or 1), the output is zero; if the inputs are different (0 and 1 or 1 and 0), the output is 1. In this case, the activation function is selected as \( f(\mathbf{Z}) = \mathbf{Z}^2 \). We assume that the input electromagnetic field is coupled to the primary mode of the waveguide, and the backscattering of the wave is neglected.

\begin{figure}[htbp]  % Specify the placement of the figure (here: 'here', 'top', 'bottom', 'page')
    \centering
    \includegraphics[width=13cm, height=3cm]{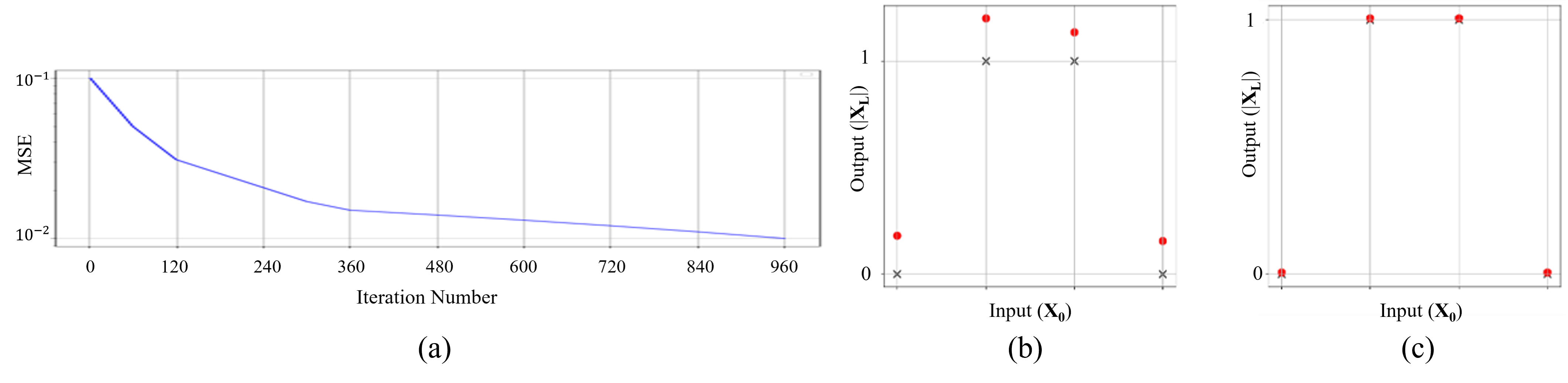}
    \caption{ONN implementing as an XOR gate: (a) Training iterations vs. MSE. (b) Initial network predictions (red circles) vs. targets (black crosses). (c) Network predictions.}
    \label{fig6}
\end{figure}

For the simulation, we used a Finite-Difference Time-Domain (FDTD) method with absorbing boundary conditions. A total of 960 iterations were considered for training this simple ONN. The result of training this model is shown in Figure \ref{fig6}.

\subsection{Optical CNN}
Using the method described in Section \ref{part3}, we design an optically equivalent CNN. In this design, we create 28 separate layers, each with 28 inputs and 28 outputs, using Mach-Zehnder Interferometer (MZI) meshes. The activation function used is ReLU.
The output performance of this model is compared to state-of-the-art models using three different algorithms: backpropagation, forward-forward, and our method, both in integrated and free-space formats. Our architecture for implementing free-space structures follows the design in \cite{r5} without changes.
These models were trained and tested on the MNIST dataset. The table below shows the test accuracy values for these models:

\begin{table}[]
\caption{\fontsize{10}{12}\selectfont Test Accuracy ($\%$) Comparison of Silica and Optical Neural Network in Integrated and Free Space Formats}
\centering
\small
\begin{tabular}{|l|c|c|c|}
\hline
\small Model                     & \small Backpropagation & \small Forward-Forward & \small Our Method \\ \hline
\small Silicon - Fully Connected             & 98.63           & 98.42           & 98.01      \\ \hline
\small Integrated - Fully Connected \cite{r12}                &91.2  &81.4  &90.2  \\ \hline
\small Free Space - Fully Connected \cite{r5}            &92.6                 &88.6                 & 91.3           \\ \Xhline{4\arrayrulewidth}
\small Silicon - CNN &99.86                 & 98.5                &98.2            \\ \hline
\small Silicon - CNN Equivalent            &99.92                 & 98.8 &98.6            \\ \hline
\small Integrated - CNN \cite{r11}           &92.9                 & 84.2                &89.9            \\ \hline
\small Integrated - CNN Equivalent           &98.54                 & 97.5                &98.1            \\ \hline
\small Free space - CNN \cite{r5}          &94.6                 & 86.4                &92.08            \\ \hline
\end{tabular}
\label{table1}
\end{table}

As can be seen in Table \ref{table1}, the proposed algorithm achieves an acceptable accuracy compared to silicon models, simple integrated ONNs, and even free-space models. This demonstrates its potential as a useful solution for processing data with larger sizes, aligning with the main goal of realizing optical neuromorphic computing in the future. Additionally, the use of this model has significantly increased the accuracy in relation to free-space optical neural networks compared to other training models.

\section{Discussion and Future Works}
In this paper, we introduced a novel approach for training ONNs using forward propagation methods inspired by biological neural processes\cite{r13}. Our method leverages modulated error signals integrated into the forward pass, eliminating the need for conventional backward propagation (backpropagation). This approach not only aligns with biological principles of neural information processing but also offers promising results in terms of computational efficiency and performance on tasks such as the MNIST dataset. In addition, the proposed algorithm generally eliminates concerns related to the design of nonlinear units in ONNs, as it does not require special design considerations or struggle with their response in both forward and backward modes.

Moreover, these approaches open new avenues for processing real-size images in integrated ONNs, addressing a challenge that has not been adequately tackled previously. The numerical results, demonstrate that our proposed optical CNN achieves competitive accuracy compared to traditional silicon-based models and existing ONNs. Particularly, our model performs well in both integrated and free-space formats, highlighting its versatility and potential for future applications in optical neuromorphic computing.
For experimental setups, the utilization of 3D inverse design structures, facilitated by advancements in 3D printing technology, could offer robust frameworks for further exploration and validation.

For future research directions, several opportunities arise from our current findings. Firstly, incorporating topological photonics phenomena to enforce unidirectional waveguides could mitigate backscattering issues, enhancing the robustness and efficiency of our optical network structure \cite{r15}. This enhancement could potentially lead to more reliable performance in real-world optical computing scenarios.
Additionally, while our current method focuses on forward propagation, future efforts could explore implementing optical backpropagation. This would necessitate the development of optical memory and computational units capable of handling complex gradient computations in reverse, akin to conventional backpropagation in neural networks. Such advancements would facilitate deeper and more complex network architectures, potentially surpassing current limitations in optical computing.

Despite these advancements, our approach has both strengths and limitations. The benefits include its biologically-inspired design, which reduces computational overhead by integrating error signals directly into the forward pass. This approach not only simplifies hardware requirements but also enhances computational speed, making it suitable for real-time applications.
However, a significant drawback lies in the current limitations of optical hardware, such as noise sensitivity and the complexity of implementing optical memory and computational units. Moreover, while our model shows competitive performance on standard datasets like MNIST, scaling up to larger and more complex datasets remains a challenge that requires further exploration.

\bibliographystyle{unsrt}
\bibliography{sample}

\begin{thebibliography}{10}

\bibitem{r1}
Giovanni Finocchio, Massimiliano Di~Ventra, Kerem~Y Camsari, Karin Everschor-Sitte, Pedram~Khalili Amiri, and Zhongming Zeng.
\newblock The promise of spintronics for unconventional computing.
\newblock {\em Journal of Magnetism and Magnetic Materials}, 521:167506, 2021.

\bibitem{r2}
Amirreza Ahmadnejad, Ahmad~Mahmmodian Darviishani, Mohmmad~Mehrdad Asadi, Sajjad Saffariyeh, Pedram Yousef, and Emad Fatemizadeh.
\newblock Tacnet: Temporal audio source counting network.
\newblock {\em arXiv preprint arXiv:2311.02369}, 2023.

\bibitem{r17}
Sina Aghili, Rasoul Alaee, Amirreza Ahmadnejad, Ehsan Mobini, Mohammadreza Mohammadpour, Carsten Rockstuhl, and Ksenia Dolgaleva.
\newblock Dynamic control of spontaneous emission using magnetized insb higher-order-mode antennas.
\newblock {\em Journal of Physics: Photonics}, 6(3):035011, 2024.

\bibitem{r3}
Yupeng Chang, Xu~Wang, Jindong Wang, Yuan Wu, Linyi Yang, Kaijie Zhu, Hao Chen, Xiaoyuan Yi, Cunxiang Wang, Yidong Wang, et~al.
\newblock A survey on evaluation of large language models.
\newblock {\em ACM Transactions on Intelligent Systems and Technology}, 15(3):1--45, 2024.

\bibitem{r4}
Kun Liao, Tianxiang Dai, Qiuchen Yan, Xiaoyong Hu, and Qihuang Gong.
\newblock Integrated photonic neural networks: Opportunities and challenges.
\newblock {\em ACS Photonics}, 10(7):2001--2010, 2023.

\bibitem{r5}
Hoda Sadeghzadeh and Somayyeh Koohi.
\newblock High-speed multi-layer convolutional neural network based on free-space optics.
\newblock {\em IEEE Photonics Journal}, 14(4):1--12, 2022.

\bibitem{r6}
Reyhane Ahmadi, Amirreza Ahmadnejad, and Somayyeh Koohi.
\newblock Free-space optical spiking neural network.
\newblock {\em arXiv preprint arXiv:2311.04558}, 2023.

\bibitem{r7}
Xianxin Guo, Thomas~D Barrett, Zhiming~M Wang, and AI~Lvovsky.
\newblock End-to-end optical backpropagation for training neural networks.
\newblock {\em arXiv preprint arXiv:1912.12256}, 2019.

\bibitem{r8}
Tyler~W Hughes, Momchil Minkov, Yu~Shi, and Shanhui Fan.
\newblock Training of photonic neural networks through in situ backpropagation and gradient measurement.
\newblock {\em Optica}, 5(7):864--871, 2018.

\bibitem{r9}
Geoffrey Hinton.
\newblock The forward-forward algorithm: Some preliminary investigations.
\newblock {\em arXiv preprint arXiv:2212.13345}, 2022.

\bibitem{r10}
Ilker Oguz, Junjie Ke, Qifei Weng, Feng Yang, Mustafa Yildirim, Niyazi~Ulas Dinc, Jih-Liang Hsieh, Christophe Moser, and Demetri Psaltis.
\newblock Forward--forward training of an optical neural network.
\newblock {\em Optics Letters}, 48(20):5249--5252, 2023.

\bibitem{r11}
Hengameh Bagherian, Scott Skirlo, Yichen Shen, Huaiyu Meng, Vladimir Ceperic, and Marin Soljacic.
\newblock On-chip optical convolutional neural networks.
\newblock {\em arXiv preprint arXiv:1808.03303}, 2018.

\bibitem{r12}
Yichen Shen, Nicholas~C Harris, Scott Skirlo, Mihika Prabhu, Tom Baehr-Jones, Michael Hochberg, Xin Sun, Shijie Zhao, Hugo Larochelle, Dirk Englund, et~al.
\newblock Deep learning with coherent nanophotonic circuits.
\newblock {\em Nature photonics}, 11(7):441--446, 2017.

\bibitem{r16}
A.~Ahmadnejad.
\newblock Training large-scale onn.
\newblock \url{https://github.com/AAhmadnejad98/Training-Large-Scale-ONN}, 2024.
\newblock Accessed: 2024-08-11.

\bibitem{r13}
G~Dellaferrera and G~Kreiman.
\newblock Error-driven input modulation: Solving the credit assignment problem without a backward pass. arxiv 2022.
\newblock {\em arXiv preprint arXiv:2201.11665}.

\bibitem{r14}
Qiang Li, Shengping Liu, Yang Zhao, Wei Wang, Ye~Tian, Junbo Feng, and Jin Guo.
\newblock Optical nonlinear activation functions based on mzi-structure for optical neural networks.
\newblock In {\em 2020 Asia Communications and Photonics Conference (ACP) and International Conference on Information Photonics and Optical Communications (IPOC)}, pages 1--3. IEEE, 2020.

\bibitem{r15}
Tianxiang Dai, Anqi Ma, Jun Mao, Yutian Ao, Xinyu Jia, Yun Zheng, Chonghao Zhai, Yan Yang, Zhihua Li, Bo~Tang, et~al.
\newblock A programmable topological photonic chip.
\newblock {\em Nature Materials}, pages 1--9, 2024.

\end{thebibliography}

\end{document}